%% file: example_paper.tex
\documentclass[nohyperref]{article}

\usepackage{microtype}
\usepackage{graphicx}
\usepackage{subfigure}
\usepackage{booktabs} 

\input{math_commands.tex}

\usepackage{hyperref}
\usepackage{url}
\usepackage{xcolor} 
\usepackage{booktabs} 
\usepackage{amsthm}
\usepackage{wrapfig}

\renewcommand{\cite}[1]{\citep{#1}}
\vbadness=\maxdimen
\definecolor{urlcolor}{rgb}{0,.145,.698}
\definecolor{mydarkblue}{rgb}{0,0.08,0.45}
\definecolor{linkcolor}{rgb}{.71,0.21,0.01}
\definecolor{citecolor}{rgb}{.12,.54,.11}
\hypersetup{ %
	pdftitle={},
	pdfauthor={},
	pdfsubject={},
	pdfkeywords={},
	pdfborder=0 0 0,
	pdfpagemode=UseNone,
	breaklinks=true,  
	colorlinks=true,
	bookmarksnumbered=true,
	urlcolor=urlcolor,
	linkcolor=linkcolor,
	citecolor=mydarkblue,
	filecolor=mydarkblue,
	pdfview=FitH}

\newcommand{\RR}{\mathbb{R}}
\newcommand{\T}{\gT}
\renewcommand{\O}{\gO}

\newcommand{\Gpoint}{\gO(3)\times S_n}

\renewcommand{\vec}{\boldsymbol}
\newcommand{\eigcov}{\lambda_{Cov}}
\newcommand{\Ffeat}{\gF_{\mathrm{feat}}}
\newcommand{\Fpool}{\gF_{\mathrm{pool}}}
\newcommand{\Wfeat}{W_{\mathrm{feat}}}

\usepackage{hyperref}

\usepackage[accepted]{icml2022}

\usepackage{amsmath}
\usepackage{amssymb}
\usepackage{mathtools}
\usepackage{amsthm}

\usepackage[capitalize,noabbrev]{cleveref}

\theoremstyle{plain}
\newtheorem{theorem}{Theorem}[section]
\newtheorem{proposition}[theorem]{Proposition}

\newtheorem{corollary}[theorem]{Corollary}
\theoremstyle{definition}

\theoremstyle{remark}

\usepackage[textsize=tiny]{todonotes}

\icmltitlerunning{A simple and universal rotation equivariant point-cloud network}

\begin{document}

\twocolumn[
\icmltitle{A simple and universal rotation equivariant point-cloud network}



\icmlsetsymbol{equal}{*}

\begin{icmlauthorlist}
\icmlauthor{Ben Finkelshtein}{technion}
\icmlauthor{Chaim Baskin}{technion}
\icmlauthor{Haggai Maron}{nvidia}
\icmlauthor{Nadav Dym}{technion}

\end{icmlauthorlist}

\icmlaffiliation{technion}{Technion -- Israel Institute of Technology}
\icmlaffiliation{nvidia}{NVIDIA Research}

\icmlcorrespondingauthor{Nadav Dym}{nadavdym@technion.ac.il}

\icmlkeywords{Machine Learning, ICML}

\vskip 0.3in
]

\printAffiliationsAndNotice{} 


\begin{abstract}

Equivariance to permutations and rigid motions is an important inductive bias for various 3D learning problems. Recently it has been shown that the equivariant Tensor Field Network architecture is universal- it can approximate any equivariant function. In this paper we suggest a much simpler architecture, prove that it enjoys the same universality guarantees and evaluate its performance on Modelnet40.
The code to reproduce our experiments is available at \url{https://github.com/simpleinvariance/UniversalNetwork}

\end{abstract}

\section{Introduction}

Permutations and rigid motions are the basic shape-preserving transformations for point cloud data. In recent years multiple neural architectures that are equivariant to these transformations were proposed, and were demonstrated to outperform non-equivariant models on a variety of 3D learning tasks, such as shape classification \cite{deng2021vector}, molecule property prediction and the $n$-body problem \cite{satorras2021n}.  

A desired and much studied 
  theoretical benchmark for equivariant neural networks is universality 
  - the ability to approximate \emph{any} continuous equivariant function. The universality of the translation equivariant convolutional neural networks was discussed in \cite{yarotsky2022universal}. In contrast, the expressive power of standard permutation-equivariant graph neural networks such as message passing networks \cite{gilmer2017neural} is limited \cite{xu2018powerful,morris2019weisfeiler,morris2021weisfeiler}. Universality for graph neural networks can be attained using a `lifted approach', where the input $n \times n$ graph is mapped to high dimensional hidden representations of the permutation group of dimension $n^{K}, K=0,1,2,\ldots $. Universality can be achieved by taking a very large $K=n$\cite{maron2019universality,ravanbakhsh2020universal}.   
  
  Our focus in this paper is on equivariant networks for $3\times n$ point clouds. In this context, universality of point cloud networks which are equivariant with respect to permutations (and not rigid motions) was discussed in \cite{zaheer2017deep,qi2017pointnet}, and the opposite case of universality with respect to rigid motions (and not permutations) was discussed in \cite{villar2021scalars,bogatskiy2020lorentz}. In most of these constructions, The size of the hidden representations in these networks is proportional to the input dimension $3\cdot n $ (though the number of channels may be large).   
  
  Many point cloud networks were suggested which are \emph{jointly} equivariant to permutations and rigid motions \cite{deng2021vector,poulenard2019effective,satorras2021n}, and have hidden representations of size $\sim 3\cdot n $. While it is currently unknown whether these networks are universal, there is reason to think they may not be, as separation of $d\times n $ point clouds up to equivalence is just as difficult as separation of graphs \cite{dym2017exact}, and this problem (known as the graph isomorphism problem) has no known polynomial time algorithm \cite{babai2016graph}.
  
  For $3\times n$ point clouds, an interesting intermediate option between the standard low dimensional approach and the untractable lifted approach is the `semi-lifted' approach. In this approach, intermediate hidden representations of dimension $3^K\cdot n, K=0,1,2\ldots $ are used, which are essentially high-dimensional representations of the rigid motion group combined with the standard representations of the permutation group. In contrast with the fully lifted approach which would require hidden representations of dimension $(3n)^K $, the complexity of the semi-lifted approach is linear in $n$ which typically is much larger than $3$. and thus for moderately sized $K$. Several papers have proposed architectures based on the semi-lifted approach, including \cite{thomas2018tensor,klicpera2021gemnet,fuchs2020se}.   
  
  Unlike the more standard low-dimensional approach, it was shown in  \cite{dym2020universality} that  semi-lifted jointly equivariant architectures  can be universal. In particular, universality can be obtained using the  Tensor Field Network (TFN) \cite{thomas2018tensor,klicpera2021gemnet,fuchs2020se} jointly equivariant layers. To the date, the construction in \cite{dym2020universality} seems to be the only universality result for jointly equivariant networks at this level of generality. Universality was obtained for the simpler cases of 2D point clouds \cite{bokman2021zz} (where the group of rotations is commutative) or 3D point clouds with distinct principal eigenvalues \cite{puny2021frame}.
 
 While universality can be obtained using TFN layers, the specific architecture used to obtain universality was not yet implemented. More importantly, TFN layers are based on the representation theory of $SO(3)$, which limits the audience of this approach and leads to cumbersome implementation.  
 
 The goal of this short paper is to present for the first time an implementation of a universal architecture using the theoretical framework laid out in \cite{dym2020universality}. This  architecture is based on tensor representations (to be defined), and as a result can be easily understood with a basic background in linear algebra. The architecture depend on hyper-parmeters $K$ and $C$ which define an architecture $\gF(K,C)$ with $C$ channels and representations of dimension at most $n\cdot 3^K $. The union over all possible $(K,C) $ is dense in the space of jointly equivariant functions.  Additionally, for fixed $K$ and large enough $C$, this architecture can express all jointly equivariant polynomials of degree $K$. In particular, we show that the eigenvalues of the point clouds's covariance matrix, which are a common invariant shape descriptor, can be expressed by architectures with $K=6$.
 
 Experimentally, with the hardware currently available to us we have been able to train models for the ModelNet40 classification class with $K\leq 6 $. These results are presented in Table~\ref{tab:ModelNet40}.    While our results are currently not state of the art, we believe this direction is worthy of further study, and may be a  good first step towards the utlimate goal of achieving simple, equivariant networks with strong theoretical properties and empirical success.

\subsection{Preliminaries}
\paragraph{Group actions and equivariance}
Given two (possibly different) vector spaces $W_1,W_2$, and a group $G$ which acts on these vector spaces, we say that $f:W_1\to W_2$ is equivariant if 
$$f(gw)=gf(w), \forall w\in W_1, g\in G .$$
 We say $f$ is \emph{invariant} in the special case where the action of $G$ on $W_2$ is trivial, that is $gw_2=w_2$ for all $g\in G$ and $w_2\in W_2$. When $G$ acts linearly on $W$ we say $W$ is a \emph{representation} of $G$.

An important principle in the design of equivariant neural networks is that they be constructed by  composition of simple equivariant functions.  To achieve models with strong expressive power, the input low dimensional representations can be equivariantly mapped `up' to high dimensional `hidden' representations such as irreducible representations \cite{thomas2018tensor,fuchs2020se} or tensor representations \cite{kondor2018covariant,maron2018invariant,maron2019universality,maron2019provably}, and them equivariantly mapped down again to the output low dimensional representation. We next introduce tensor representations which will be used in this paper.      

\paragraph{Tensor representations}
Let $\T_k$ denote the vector spaces $\T_0=\RR, \, \T_1=\RR^3, \, \T_2=\RR^{3\times 3}, \, \T_3=\RR^{3\times 3 \times 3}, \ldots \quad $. An orthgonal matrix $R\in \O(3)$ acts on $\T_k$ via
$$\left(R^{\otimes k}V \right)_{i_1,\ldots,i_k}=\sum_{j_1,\ldots,j_k=1}^3 R_{i_1,j_1}R_{i_2,j_2}\ldots R_{i_k,j_k}V_{j_1,j_2,\ldots,j_k}. $$
We note that  $R^{\otimes k}:\T_k \to \T_k$ can be identified with a mapping  $R^{\otimes k}:\RR^{3^k} \to \RR^{3^k}$, and as our notation suggests this mapping is the Kronecker product of $R$ with itself $k$ times. 
For $k=0,1,2$ applying $R^{\otimes k}$ to the scalar/vector/matrix $V\in \T_k$ gives
$$R^{\otimes 0} V=V, \quad R^{\otimes 1} V=RV \text{ and } \quad R^{\otimes 2} V=RVR^T .  $$

\textbf{Equivariant mappings} we review two basic basic mappings between tensor representations of $\gO(3)$, which will later be used to define our equivariant layers.

\emph{Tensor product} mappings are a standard method to equivariantly map lower order representations to higher order representations. 
The tensor product $\otimes:\T_{k} \times \T_{\ell} \to \T_{k+\ell}  $ is defined for $V^{(1)}\in \T_k   $ and $V^{(2)}\in \T_\ell   $ by 
$$\left(V^{(1)} \otimes V^{(2)}\right)_{i_1,\ldots,i_k,j_1,\ldots,j_\ell}=V^{(1)}_{i_1,\ldots,i_k}\cdot V^{(2)}_{j_1,\ldots,j_\ell}. $$

 A proof of the equivariance of  tensor product is given in Proposition~\ref{prop:equi1} in the appendix.
For now we give a simple example: when $k=\ell=1$ the equivariance   of the tensor product  mapping can be seen by noting that  for vectors $v,w\in \T_1$ and $R\in \O(3) $ we obtain 
$$(Rv) \otimes (Rw)=(Rv)(Rw)^T=Rvw^TR^T=R^{\otimes 2}(v\otimes w).$$

\emph{Contractions} are a convenient method for equivariantly mapping high order representations to lower order representations: 
for $k\geq 2$, any pair of indices $a,b$ with $1\leq a<b\leq k$ defines a contraction mapping $C_{a,b}:\T_k \to \T_{k-2}$, which is defined by jointly marginalizing  over the $a,b$ indices. For example for 
 $a=2,b=k$ we have
$$\left( C_{2,k}(V) \right)_{i_1,i_2,\ldots,i_{k-2}}=\sum_{j=1}^3 V_{i_1,j,i_2\ldots,i_{k-2},j} .$$
 The equivariance of  contractions will be proved in Proposition~\ref{prop:equi1} in the appendix. Fir now we present a simple example is the case $k=2,a=1,b=2$ where we get for every matrix $V\in \T_2$ that  
$$C_{1,2}(V)=\sum_{j=1}^3 V_{jj}=\mathrm{trace}(V) $$
and $\mathrm{trace}:\T_2\to \T_0 $ is indeed a $\gO(3)$ equivariant mapping.

\section{Method}

\subsection{Setup and overview}
Our goal is to construct an architecture which produces equivariant functions $f:W_0\to W_1$, where $W_0=\RR^{3 \times n}$ is the space of point clouds, which is acted on by the group of orthogonal transformations and permutations $\O(3)\times S_n$. We note that translation equivariance/invariance can be easily added to our model by centralizing the input point cloud to have zero mean (see \cite{dym2020universality} for more details). The output representations $W_1$ we consider vary according to the task at hand: for example, classification tasks are invariant to rigid motions and permutations, predicting the trajectory of a dynamical system is typically equivariant to permutations and rigid motion, while segmentation tasks are permutation equivariant but invariant to rigid motions.

As in previous works our architecture is a concatenation of several equivariant layers that will be discussed in detail next. Specifically, it is composed of three main types of  layers: \emph{Ascending Layers} that produce higher order representations, \emph{Descending Layers} which do the opposite and Linear layers that allow us to mix different channels. We use a U-net based \cite{ronneberger2015u} architecture that first uses ascending layers up to some predefined maximal order $K$, and then descends to the required output order (see Figure \ref{fig:arch}).

\subsection{Equivariant layers}
In general we consider mappings between representations of $\Gpoint$
of the form $\T_k^{n \times C} $ where the group action is given by 
\begin{equation}\label{eq:joint_action}
\left[(R,\sigma)(V)\right]_{j,c}=R^{\otimes k}(V_{\sigma^{-1}(j),c}) 
\end{equation}
where we denote elements  in $\T_k^{n \times C}$ by $V=(V_{jc})_{1\leq j \leq n, 1\leq c\leq C}$
and $V_{jc}\in \T_k $ for every fixed $j,c$. 
Note that for $k=1,C=1$ we get our input representation $\T_1^{n \times 1}=\RR^{3\times n} $.  Our construction is based on three basic layers:

\textbf{Ascending Layers:}
Ascending layers are parametric mappings $\gA:\T_k^{n \times C} \times \RR^{3 \times n} \to \T_{k+1}^{n \times C}$ which depend on parameters $\vec{\alpha}=(\alpha_{1c},\alpha_{2c})_{ c=1,\ldots,C}$. They use tensor products to obtain higher order represenations and  are  of the form\footnote{Note that this layer was already suggested in \cite{dym2020universality}, and its structure resembles the structure of the basic layers in \cite{zaheer2017deep,maron2020learning,thomas2018tensor} }
$V^{out}=\gA(V^{in},X|\vec{\alpha}) $,
where (using $X_j$ to denote the $j$-th column of $X$)
\begin{align}\label{eq:ascending}
    V^{out}_{jc}&= \alpha_{1c}\left(X_j \otimes V_{jc}\right) + \alpha_{2c}\sum_{i\neq j} X_i \otimes V_{ic}
\end{align}

\textbf{Descending layers:}
Descending layers are parameteric mappings $\gD:\T_k^{n\times C} \to \T_{k-2}^{n\times C}  $ (defined for $k\geq 2$) which   are of the form $V^{out}=\gD(V^{in}|\vec{\beta}) \text{, where } \vec{\beta}=(\beta_{a,b,c})_{1\leq a<b\leq k, 1\leq c \leq C} . $
and are defined by:
$$V^{out}_{j,c}=\sum_{1 \leq a<b \leq k} \beta_{a,b,c}C_{a,b}(V^{in}_{j,c})$$

\textbf{Linear layers:} We use the linear layers from \cite{thomas2018tensor}. These layers are parametric mappings  $\gL:\T_k^{n \times C} \to \T_k^{n \times C'}  $ of the form 
$V^{out}=\gL(V^{in}|\vec{\gamma}) \text{, where }   \vec{\gamma}=(\gamma_{cc'})_{1\leq c \leq C, 1\leq c' \leq C' }$,
which are defined by:
$$V^{out}_{jc'}=\sum_{c=1}^C \gamma_{cc'}V_{jc}^{in}. $$ 
The equivariance of these layers to orthogonal transformations and permutations follows rather easily from our previous discussion. We prove this formally in Proposition~\ref{prop:equi2} in the appendix.

\subsection{Architecture} 

\begin{figure}[t]
\includegraphics[width=\columnwidth]{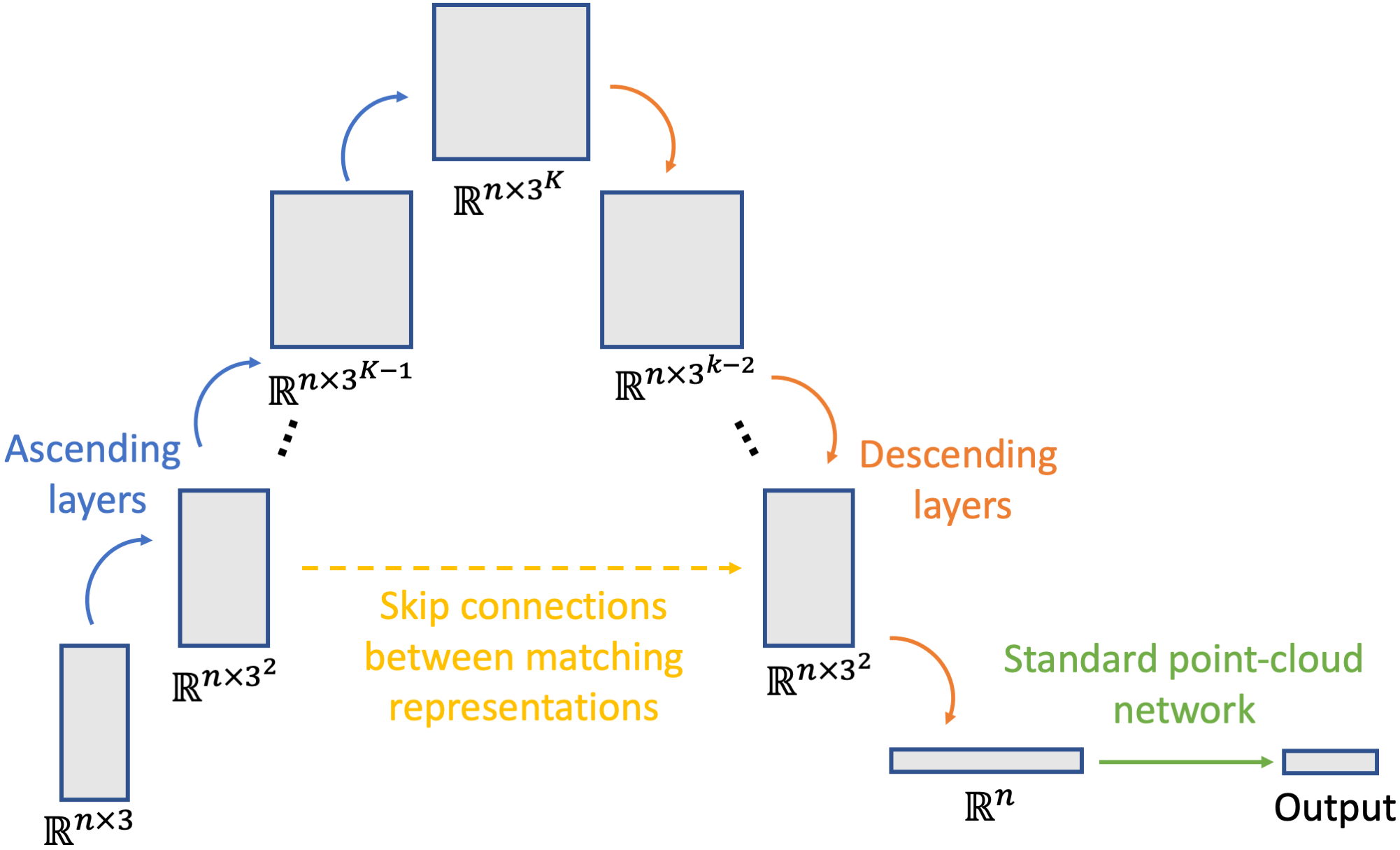}
\caption{Our architecture $\gF(K,C)$ with $C=1 $.}
\label{fig:arch}
\end{figure}

The architecture we use depends on two hyper-parameters: the number of channels $C$ (which we keep fixed throughout the network), and the maximal representation order used $K$. This choice of hyper-parameters defines a parametric function space $\gF(K,C)$, containing functions which gradually map pointclouds up to $K$ order representations, and then gradually map back down, using the ascending, descending and linear layers discussed above. The architecture is visualized in Figure~\ref{fig:arch} (with $C=1$ and using the identification $\T_k=\RR^{3^k}$). We next formally describe our architecture.

We denote  the input point cloud by $X\in \RR^{3 \times n}$, and define an initial degenerate representation $U^{(0)}\in \T_0^{n \times C}$ which is identically one in all $n\cdot C$ coordinates. we recursively define for $k=1,\ldots,K$ 
$$U^{(k)}=\gL \left( \gA(U^{(k-1)},X)\right) .$$
where we suppress the dependence of $\gA$ and $\gL$ on the learned parameters $\vec{\alpha}^{(k)} $ and $\vec{\gamma}^{(k)} $  for notation simplicity. Each $U^{(k)}=U^{(k)}(X)$ contains $n\times C$ copies of a $3^k$ dimensional tensor in $\T_k$. Next we denote $U^{(K)}=V^{(K)}$ and recursively define for $k=K,K-2,\ldots,r+2$ where $r=K (\mod 2)$
\begin{equation}\label{eq:U}
V^{(k-2)}=\gL \left( concat(\gD(V^{(k)}),U^{(k-2)})\right)  \end{equation}
where again we suppress the dependence of $\gD$ and $\gL$ on the learned parameters $\vec{\beta}^{(k-2)} $ and $\bar{\vec{\gamma}}^{(k-2)} $  for simplicity.
Overall we get a function $f:\T_1^{n\times C}\to \T_r^{n \times C} $ of the form
\begin{align*}
V^{(r)}=f(X|\vec{\alpha}^{(1)},\ldots&,\vec{\alpha}^{(K)},\vec{\beta}^{(K-2)},\ldots,\vec{\beta}^{(r)},\\
&\vec{\gamma}^{(1)},\ldots,\vec{\gamma}^{(K)},\bar{\vec{\gamma}}^{(K-2)},\ldots,\bar{\vec{\gamma}}^{(r)}).
\end{align*}
The function $f$ is  permutation and ortho-equivariant when $r=1$ or ortho-\emph{invariant} and permutation equivariant when $r=0$. When $r=0$ we can apply a permutation invariant/equivariant network such as PointNet \citep{qi2017pointnet} or DGCNN \citep{wang2019dynamic} to the output of our network to strengthen the expressive power of our network while maintaining the  ortho-invariance and permutation invariance/equivariance of our overall construction. 

\paragraph{Expansion of basic model}
The basic model we presented up to now is sufficient to prove universality as discussed in Theorem~\ref{thm:universal} below. However, it is \emph{only} able to compute polynomials up to degree $K$. To enable expression of non-polynomial functions we use the ReLU activation layer defined in \cite{deng2021vector} (which easily generalizes to higher representations). To take local information into account, we expand our ascending layer in \eqref{eq:ascending} by adding a summation over the K-nearest neighbors in feature space, giving \begin{align*}
    V^{out}_{jc}&= \alpha_{1c}\left(X_j \otimes V_{jc}\right) + \alpha_{2c}\sum_{i\neq j} X_i \otimes V_{ic}\\
    &+\alpha_{3c}\sum_{i\sim j} X_i \otimes V_{ic}.
\end{align*}
Finally, we have experimented with adding linear layers which map $\T_k$ equivariantly to itself. When $k=2$ these linear layers are spanned by the linear mappings
$$V\mapsto V, \, V\mapsto V^T, \, V \mapsto trace(V)I_3, \quad V\in \T_2=\RR^{3\times 3} .$$
We find that adding these mappings for $k=2$ improves our results, and generalizing this to higher dimensions is an interesting challenge for further research.


\section{Theoretical properties}
We now discuss the expressive power of the architecture $\gF(K,C)$ defined above. Based on the proof methodology of \cite{dym2020universality}, we prove the following theorem (stated formally in the appendix)
\begin{theorem}\label{thm:universal}[non-formal statement]
For every even $K$ and large enough $C\geq C(K)$, every polynomial of degree $\leq K$ which is permutation equivariant (or invariant) and invariant to rigid motions can be expressed by functions in  $\gF(K,C)$ composed with simple pooling and centralizing operations.
\end{theorem}

Since the invariant/equivariant polynomials are dense in the space of continuous invariant functions (uniformly on compact sets, see e.g., Lemma 1 in \cite{dym2020universality}) this theorem means that in the limit where $K,C\rightarrow \infty$ our architecture is able to approximate any function invariant to orthogonal transformations, translations and permutations. 

A disadvantage of universality theorems is that it is unclear how big a network is needed to get a reasonable approximation of a given function. In Theorem~\ref{thm:eigs}, stated and proved in the appendix, we consider the  function $\eigcov$ which computes the three eigenvalues of the covariance matrix of the point cloud $X$, ordered according to size. This is a classical global descriptor (see e.g.,\cite{puny2021frame,kazhdan2004shape}) which  is invariant to permutations and rigid motions. We show that it can be computed by our networks with representations of order $K=6$, composed with a continuous (non-invariant) function $q$ which can be approximated by an MLP.

\section{Experiments}
\begin{table*}[hbt!]
\centering
    \small
    \vspace{3em}
    \begin{tabular}{lrrr}
    \toprule
    \textbf{Methods} &
    \multicolumn{1}{l}{z/z}	& \multicolumn{1}{l}{z/SO(3)} & \multicolumn{1}{l}{SO(3)/SO(3)}\\
    \midrule
    SFCNN \citet{rao2019spherical} & 91.4  & 84.8  & 90.1 \\
    TFN \citet{thomas2018tensor}  & 88.5  & 85.3  & 87.6 \\
    RI-Conv \citet{zhang2019rotation} & 86.5  & 86.4  & 86.4 \\
    SPHNet \citet{poulenard2019effective} & 87.7  & 86.6  & 87.6 \\
    ClusterNet \citet{8954468} & 87.1  & 87.1  & 87.1 \\
    GC-Conv \citet{zhang2020global} & 89.0    & 89.1  & 89.2 \\
    RI-Framework \citet{2021}& 89.4  & 89.4  & 89.3 \\
    VN-PointNet \citet{deng2021vector} & 77.5  & 77.5  & 77.2 \\
    VN-DGCNN \citet{deng2021vector} & 89.5  & 89.5  & 90.2 \\
    \midrule
    Our method (K=2) & 78.3 & 78.4 & 77.7 \\
    Our method (K=4) & 80.4 & 80.4 & 78.8 \\
    Our method (K=6) & 81.6 & 81.5 & 80.1\\
    \midrule
    Ours (K=4) w.o. VNReLU (K=4) & 51.4 & 51.4 & 55.2\\
    Ours (K=4) w.o. VNReLU \& KNN (K=4) & 35.8 & 35.8 & 33.8\\
    \bottomrule
    \end{tabular}%
    \caption{Test clasification accuracy on the ModelNet40 dataset in three train/test scenarios. z stands for the aligned data augmented by random rotations around the vertical axis and SO(3) indicates data augmented by random rotations. All models presented in the table are rotation and permutation invariant (up to numerical inaccuracies).}
  \label{tab:ModelNet40}%
\end{table*}
In this section, we evaluate our model. In \cref{subsec:experimental_setting}, we describe in details the experimental setup we have used and in \cref{subsec:main_results} we present our main results.

\subsection{Experimental Setting}
\label{subsec:experimental_setting}
\paragraph{Data} We evaluated our proposed network on a standard point cloud classification benchmark, ModelNet40 \cite{wu20153d}, which consists of 40 classes with 12,311 pre-aligned CAD models, split into  80\% for training and 20\% for testing. Classification problems are invariant to rigid motions and permutations. We preprocess the point cloud to have zero mean to achieve translation invariance, and then apply  the model described in the paper with $K$ even to achieve function invariant to rigid motions and equivariant to permutations. Finally we apply sum pooling and a fully connected neural network to achieve a fully invariant model.

\paragraph{Implementation} We trained and tested our model on NVidia GTX A6000 GPUs with python 3.9, Cuda 11.3, PyTorch 1.10.0, PyTorch geometric, and pytorch3d 0.6.0. We trained our model for 100 epochs with a batch size of 32, a learning rate of 0.1, and a seed of 0.

\subsection{Experimental Results}
\label{subsec:main_results}
We present initial results of our model on the ModelNet40 classification task, under the three standard  evaluation protocols for jointly equivariant architectures: in the first protocol (z/z)  train and test data are augmented with rotations around the z-axis, in the second (z/SO(3)) train data is augmented by rotations around the z-axis and test data is augmented by general 3D rotations, and in the third protocol (SO(3)/SO(3)) train and test data are augmented by 3D rotations.  

We compared our method to SFCNN \cite{rao2019spherical}, TFN \cite{thomas2018tensor}, RI-Conv \cite{zhang2019rotation}, SPHNet \cite{poulenard2019effective}, ClusterNet \cite{8954468}, GC-Conv \cite{8954468}, RI-Framework \cite{2021} and to two variations of Vector Neurons \cite{deng2021vector}. The results of the comparison are shown in Table~\ref{tab:ModelNet40}. In general We find that our model does not perform as well as many of the models we compared too.

All methods we compared to, as well as our own method, are invariant to rigid motions and permutations, and as a result are hardly effected when the test data is augmented by general 3D rotations (z/SO(3)) or when both test and train are augmented by 3D rotations (SO(3)/SO(3)) (for comparison, see the results for the three augemntation protocols obtained for methods without rotation invariance, as shown  e.g., in \cite{deng2021vector}). 

\paragraph{Ablation on our method} We examine the contribution of high representations by comparing different representation orders $K=2,4,6$. As expected we find that higher representation lead to better accuracy. Without the use of the ReLU activation layer defined in \cite{deng2021vector}, the accuracy of our method decreases by 29.0\%. When we remove the K-nearest neighbors summation as well, the accuracy of our method decreases by an additional 15.6\%.

\section{Conclusion and future work} In this short paper we first presented an implementation of a simple, universal, neural network architecture which is jointly equivariant of permutations and rigid motions. Experimentally, we find that our model does not perform as well as  recent state of the art architectures with joint permutation and rigid motion invariance. We are currently working on additional expansions to our basic model, such as the study of equivariant linear mappings on $\T_k$ mentioned above, and believe this may lead to improved results on Modelnet40 and other equivariant learning tasks. 

At the same time, we believe the inability of our method, as well as other semi-lifted approaches such as TFN, to outperform methods with low-dimensional representations is worthy of further study and raises many interesting questions for further reseach. For example: Are the low dimensional models universal? If they aren't, what equivariants tasks are they likely to fail in? Is there an explanation to their successfulness in other tasks? Is it possible that semi-lifted models are more expressive but are difficult to optimize? If so can this difficulty be alleviated? We hope to address these questions in future work, and hope others will be inspired to do the same.    


\newpage

\bibliography{example_paper}
\bibliographystyle{icml2022}
\clearpage
\onecolumn
\appendix

\section{Equivariance proofs}
We begin by proving
\begin{proposition}\label{prop:equi1}
Tensor products and contractions are $\gO(3)$ equivariant. 
\end{proposition}
As a preliminary to this proof, recall that $T\in \T_k$ is a rank one tensor if there exists $t^{(1)},\ldots,t^{(k)}\in \RR^3$ such that 
\begin{equation}\label{eq:rank_one}
T_k=t^{(1)}\otimes t^{(2)}\otimes \ldots \otimes t^{(k)} .
\end{equation}
\begin{proof}[Proof of Proposition~\ref{prop:equi1}]
\emph{equivariance of tensor products}  

We need to show that 
for $T\in \T_k   $ and $S\in \T_\ell   $ we have 
\begin{equation}\label{eq:tens_equi}
R^{\otimes k}T \otimes R^{\otimes \ell}S= R^{\otimes (k+\ell)} \left( T \otimes S \right), \quad \forall R \in \gO(3) .
\end{equation}
since both sides of the equation are bilinear in $(T,S)$, and since every tensor in $\T_k$ can be written as a linear combination of rank one tensors, it is sufficient to prove \eqref{eq:tens_equi} for the special case where $T$ and $S$ are rank one tensors. If $T$ is a rank one tensor as in \eqref{eq:rank_one}, then $R^{\otimes k}T $ is given by 
\begin{align*}
\left[(Rt^{(1)})\otimes (Rt^{(2)}) \otimes  \ldots \otimes (Rt^{(k)}) \right]_{i_1,\ldots,i_k}
&= (Rt^{(1)})_{i_1}\times (Rt^{(2)})_{i_2} \times  \ldots \times (Rt^{(k)})_{i_k}\\
&=\left(\sum_{j_1=1}^3 R_{i_1j_1}t_{j_1}^{(1)}\right)\left(\sum_{j_2=1}^3 R_{i_2j_2}t_{j_2}^{(2)}\right)\ldots\left(\sum_{j_k=1}^3 R_{i_kj_k}t_{j_k}^{(k)}\right)\\
&=\sum_{j_1,\ldots,j_k=1}^3 R_{i_1,j_1}R_{i_2,j_2}\ldots R_{i_k,j_k}T_{j_1,j_2,\ldots,j_k}\\
&=\left(R^{\otimes k}T \right)_{i_1,\ldots,i_k}
\end{align*}
it follows that for every rank one tensors $T=t^{(1)}\otimes t^{(2)} \otimes \ldots \otimes t^{(k)} $ and $S=s^{(1)}\otimes s^{(2)} \otimes \ldots \otimes s^{(\ell)} $ we have for every $R\in \gO(3)$
\begin{align*}
R^{\otimes k}T \otimes R^{\otimes \ell}S &=(Rt^{(1)})\otimes (Rt^{(2)}) \otimes  \ldots \otimes (Rt^{(k)})\otimes (Rs^{(1)})\otimes (Rs^{(2)}) \otimes  \ldots \otimes (Rs^{(\ell)})\\
&=R^{\otimes(k+\ell)}\left[t^{(1)}\otimes t^{(2)}\otimes \ldots \otimes t^{(k)} \otimes s^{(1)} \otimes s^{(2)} \otimes \ldots \otimes s^{(\ell)}\right]\\
&=R^{\otimes(k+\ell)}(T \otimes S)
\end{align*}
and thus we have shown correctness of \eqref{eq:tens_equi} for all rank-one tensors and thus for all tensors. We note that this proof does not really require $R$ to be an orthogonal matrix and would work for any square matrix. This is not the case for contractions, as we will see next:

\emph{equivariance of contractions}
For simplicity of notation we prove the equivariance of  contractions $C_{a,b}:\T_k \to \T_{k-2}$ in the special case $a=k-1,b=k$. We need to show that for all $T\in \T_k $ and $R\in \gO(3) $ we have 
\begin{equation}\label{eq:contraction_equi}
C_{k-1,k}(R^{\otimes k} T)=R^{\otimes (k-2)}C_{k-1,k}( T)
\end{equation}
since both sides of the equation above are linear in $T$ and every tensor of order $k$ can be written as a linear combination of rank one tensors, it is sufficient to show that 
\eqref{eq:contraction_equi} holds for all rank one tensors. Let $T$ be a rank one tensor as in \eqref{eq:rank_one}. Note that by definition
\begin{align}\label{eq:cont_ip}
\left[C_{k-1,k}(T)\right]_{i_1,i_2,\ldots,i_{k-2}}&= \sum_{j=1}^3 T_{i_1,\ldots,i_{k-2},j,j}\\
&=\langle t^{(k-1)},t^{(k)}\rangle \left[ t^{(1)}\otimes t^{(2)} \otimes \ldots \otimes t^{(k-2)} \right] \nonumber
\end{align}
and so for every $R\in \gO(3)$, since we saw in the first part of the proof that $R^{\otimes k}T$ is a rank one tensor given by 
$$R^{\otimes k}T=(Rt^{(1)})\otimes (Rt^{(2)}) \otimes  \ldots \otimes (Rt^{(k)})$$
and so we get that 
\begin{align*}
 C_{k-1,k}(R^{\otimes k}T)&=\langle Rt^{(k-1)},Rt^{(k)}\rangle (Rt^{(1)})\otimes (Rt^{(2)}) \otimes \ldots (Rt^{(k)})\\
 &=\langle t^{(k-1)},t^{(k)}\rangle (Rt^{(1)})\otimes (Rt^{(2)}) \otimes \ldots (Rt^{(k)})\\
 &=\langle t^{(k-1)},t^{(k)}\rangle\left[R^{\otimes(k-2)}\left(t^{(1)}\otimes t^{(2)} \otimes \ldots \otimes t^{(k-2)}\right) \right]\\
 &=R^{\otimes (k-2)}C_{k-1,k}( T).
\end{align*}
Thus we have shown that \eqref{eq:contraction_equi} holds for all rank one tensors and thus for all tensors. This concludes the prove of Proposition~\ref{prop:equi1}.
\end{proof}

\textbf{Layer equivariance}
We now prove equivariance of our layers. 
In the following discussion we use the notation $\rho_k(R,\sigma)$ for the action of $(R,\sigma)$ on $\T_k^{n\times C}$ as defined in \eqref{eq:joint_action}.
\begin{proposition}\label{prop:equi2}
For any choice of parameters, the layers $\gA,\gD$ and $\gL$ are equivariant.
\end{proposition}

\begin{proof}[Proof of Proposition~\ref{prop:equi2}]

\textbf{Equivariance of ascending layers} We need to show that for every given input $X\in \RR^{3\times n}, V\in \T_k^{n \times C}$, for every $R\in \gO(3), \sigma \in S_n$ and any fixed parameter vector $\vec{\alpha}$
$$\rho_{k+1}(R,\sigma)V^{out}=\gA(\rho_{k}(R,\sigma)V^{in},\rho_{1}(R,\sigma)X|\vec{\alpha}) $$
Indeed using the definition of the action $\rho_{k+1}$ from \eqref{eq:joint_action} and the equivariance of the tensor prouct we proved above, we have 
\begin{align*}\left[\rho_{k+1}(R,\sigma)V^{out} \right]_{jc}&=R^{\otimes (k+1)}[V^{out}_{\sigma^{-1}(j),c}]\\
&=\alpha_{1c} R^{\otimes (k+1)}\left(X_{\sigma^{-1}(j)} \otimes V_{\sigma^{-1}(j)c}\right) + \alpha_{2c}\sum_{i\neq \sigma^{-1}(j)} R^{\otimes (k+1)} \left( X_i \otimes V_{ic} \right)\\
&=\alpha_{1c} (RX_{\sigma^{-1}(j)}) \otimes (R^{\otimes k}V_{\sigma^{-1}(j)c}) + \alpha_{2c}\sum_{i\neq \sigma^{-1}(j)}  (R X_i) \otimes (R^{\otimes k} V_{ic} )\\
&=[\gA(\rho_{k}(R,\sigma)V^{in},\rho_{1}(R,\sigma)X|\vec{\alpha})]_{j,c}
\end{align*}

\textbf{Equivariance of descending layers} We need to show that for all $R\in \gO(3), \sigma\in S_n$, for all $ V^{in} \in \T_k^{n \times C}, V^{out} \in \T_{k-2}^{n \times C}$ and for all choice of a parameter vector $\vec{\beta}$, 
$$\rho_{k-2}(R,\sigma)V^{out}=\gD(\rho_k(R,\sigma)V^{in}|\vec{\beta}).$$
Using the definition of the action $\rho_{k}$ from \eqref{eq:joint_action} and the equivariance of contraction we proved above, we  have
\begin{align*} 
[\rho_{k-2}(V^{out})]_{j,c}&= R^{\otimes (k-2) }[V^{out}_{\sigma^{-1}(j),c}]=\sum_{1 \leq a<b \leq k} \beta_{a,b,c}R^{\otimes (k-2)}C_{a,b}(V^{in}_{\sigma^{-1}(j),c})\\
&=\sum_{1 \leq a<b \leq k} \beta_{a,b,c}C_{a,b}(R^{\otimes k}V^{in}_{\sigma^{-1}(j),c})=[\gD(\rho_k(R,\sigma)V^{in}|\vec{\beta})]_{j,c}
\end{align*}

\paragraph{Equivariance of linear layers} We need to show that for all $R\in \gO(3), \sigma\in S_n$, for all $ V^{in}, V^{out} \in \T_{k}^{n \times C}$ and for all choice of a parameter vector $\vec{\gamma}$, 
$$\rho_{k}(R,\sigma)V^{out}=\gL(\rho_{k}(R,\sigma)V^{in}|\vec{\gamma}).$$
Indeed
\begin{align*}
[\rho_{k}(R,\sigma)V^{out}]_{jc'}&=R^{\otimes k}[V^{out}_{\sigma^{-1}(j),c'}]=\sum_{c=1}^C \gamma_{cc'}R^{\otimes k}V_{\sigma^{-1}(j)c}^{in} \\
&=[\gL(\rho_{k}(R,\sigma)V^{in}|\vec{\gamma})]_{jc'}.
\end{align*} 
This concludes the prove of Proposition~\ref{prop:equi2}.
\end{proof}

\section{Expressive power proofs}
We now reformulate and prove Theorem~\ref{thm:universal}.  \begin{theorem}[Reformulation of Theorem~\ref{thm:universal}]
For every even number $K$ and every large enough $C=C(K)$, every polynomial $p:\RR^{3\times n}\to \RR^n$ of degree $\leq K$ which is permutation equivariant and invariant to rigid motions can be obtained as the first channel of a function $f\in \gF(K,C)$, that is 
$$p_j(X)=f_{j1}(X), \forall j=1,\ldots,n $$
\end{theorem}
This immediately implies an analogous result where we replace the permutation equivariant assumption by a permutation invariant assumption:
\begin{corollary}
For every even number $K$ and every large enough $C=C(K)$, every polynomial $p:\RR^{3\times n}\to \RR$ of degree $\leq K$ which is \emph{invariant} to permutations and rigid motions can be obtained by applying sum pooling to the first channel of a function $f\in \gF(K,C)$, that is 
$$p(X)=\sum_{j=1}^nf_{j1}(X-\frac{1}{n}X1_n1_n^T)  $$
\end{corollary}

\begin{proof}[proof of Theorem~\ref{thm:universal}]
Our proof is based on the general framework for proving universality laid out in \cite{dym2020universality}. In this paper it is shown that polynomials $p:\RR^{3 \times n} \to \RR^n$ of degree $K$ which are permutation equivariant and  invariant to translations and orthogonal transformations\footnotemark can be written for large enough $C$ as
\begin{equation}\label{eq:rep}
p(X)=\sum_{c=1}^C \hat{\Lambda}_c\left(g_c(X)\right) 
\end{equation}
where 
\begin{enumerate}
\item $g_c$ is a member of a function space $\Ffeat$ which maps $\RR^{3\times n}$ to $\Wfeat^n$, where $\Wfeat$ is a representation of $\gO(3)$.

\item $\Lambda_c$ is a member of a space of  functions $\Fpool $ from $\Wfeat$ to $\RR$ and $\hat{\Lambda}_c:\Wfeat^n \to \RR^n$ is the function induced by elementwise application of $\Lambda_c$.

\item The function spaces $\Fpool$ has the linear universality property. This means that $\Fpool$ is precisely the set of linear functionals $\Lambda:\Wfeat \to \RR$ which are $\O(3)$ equivariant.  

\item the function space $\Ffeat$ has the $K$-spanning property. This means that all functions in $\Ffeat$ are required to by $\gO(3)\times S_n$ equivaraint and invariant to translations, and additionally, that any permutation equivariant and translation invariant (ortho-equivariance not required) degree $D$ polynomial $\tilde{p}:\RR^{3\times n}\to \RR^n$  can be obtained by an expression as in \eqref{eq:rep} where $\Lambda_c$ are linear but are not required to be ortho-invariant. 
\end{enumerate}

\footnotetext{Actually the argument in this paper discusses  rotation invariance rather than orthogonal transformations (that is, we also discuss invariance to reflections). However the arguments there hold in this case as well.}

It is also shown in Lemma 3 in \cite{dym2020universality} that the spaces of functions obtained by applying  ascending layers recursively $k=1,\ldots,K$ times form a $K$-spanning family. Here we use ascending layers independently on each channel and essentially remove the linear layers by setting   $\vec{\gamma}^{(k)}_{cc'}=\delta_{cc'}$ for all $k=1,\ldots,K$. Applying ascending layers $k$ times gives us a function from $\RR^{3\times n}$ to $\T_k$, and since we are basically considering a collection of different functions to different representations $\T_1,\ldots,\T_K$ (depending on the number of ascending layers used) we choose to embed all these representations into a joint representation $\Wfeat=\oplus_{k=1}^K \T_k $. Thus we see that every permutation equivaraint and translation and ortho-invariant polynomial $p$ can be written in the form \eqref{eq:rep} where the $g_c$ are obtained by applying our ascending layers $k_c$ times for some $1\leq k_c \leq K$ and $\Lambda_c:\oplus_{k=1}^K \T_k \to \RR $ is linear and ortho-invariant. Since $g_c$ maps into a single representation $\T_{k_c}$ in practice we can think of  $\Lambda_c$ is a linear equivariant functional on this single representation. 

Using an analogous argument to the proof of Proposition 1 in Appendix 5 in \cite{dym2020universality}, it can be shown that the space of linear invariant functionals $\Lambda:\T_k\to \RR$ are spanned by the functionals $\{\Lambda_\sigma| \, \sigma \in S_k\}$ which are defined uniquely be the requirement that for every $x^{(1)},\ldots,x^{(k)}\in \RR^3$,
$$\Lambda_\sigma(x^{\sigma(1)})\otimes \ldots \otimes x^{\sigma(k)})=\langle x^{\sigma(1)},x^{\sigma(2)}\rangle \times \langle x^{\sigma(3)},x^{\sigma(4)}\rangle \ldots \langle x^{\sigma(k-1)},x^{\sigma(k)}\rangle .  $$
These $\Lambda_\sigma$ are given by (see also the derivation of  \eqref{eq:cont_ip})
$$\Lambda_\sigma(T)=C_{\sigma(1),\sigma(2)}\circ C_{\sigma(3),\sigma(4)}\circ \ldots \circ C_{\sigma(k-1),\sigma(k)}(T) . $$
In particular we see that if $k$ is odd there is no non-zero linear equivariant functional from $\T_k$ to $\RR$, so we can assume that $k_c$ is even for all $c=1,\ldots,C$, and $\Lambda_c$ can be obtained by applying the last $k_c/2$ descending layers of our construction to the output $U^{(k_c)} $ of $g_c$ with an appropriate choice of $0-1$ parameters. Recall that $U^{(k_c)}=g_c(X) $  was obtained from $k_c$ ascending layers. When $k_c<K$ we achieve this using our $K$-dimensional $U$-shaped architecture  by `short-circuiting'  at the $k_c$ level, that is by setting the parameters of the linear layer $\gL$ in \eqref{eq:U} to erase $\gD(U^{(k_c+2)}) $ and maintain only $U^{(k_c)} $.

We have seen that we can write $p$ as in \eqref{eq:rep} where $\hat \Lambda_c$ and $g_c$ can be constructed to be the $c$-th channel of  our network. We obtain the sum of $\hat \Lambda_c \circ g_c$ in the first channel of the output representation $V^{(0)}$ by setting the first column of the parameter matrix $\vec{\gamma}^{(0)}$ of the last linear layer\\
to be $\gamma_{c1}^{(0)}=1 , \forall c=1,\ldots,C$ .

\end{proof}

\textbf{Computing eigenvalues of the covariance matrix} We prove 
\begin{theorem}\label{thm:eigs}
Let $\eigcov:\RR^{3\times n}\to \RR^3$ be the function which, given a point cloud $X\in \RR^{3\times n}$, computes the ordered eigenvalues of the covariance matrix $(X-\frac{1}{n}X1_n1_n^T)(X-\frac{1}{n}X1_n1_n^T)^T $.
For large enough $C$ there exists $f\in \gF(6,C)$ and a continuous $q:\RR^3\to \RR^3$ such that 
$$\eigcov(X)=q\left(\sum_{j=1}^nf_{j1}(X-\frac{1}{n}X1_n1_n^T),\sum_{j=1}^nf_{j2}(X-\frac{1}{n}X1_n1_n^T),\sum_{j=1}^n f_{j3}(X-\frac{1}{n}X1_n1_n^T)\right) $$
\end{theorem}

\begin{proof}[proof of Theorem~\ref{thm:eigs}]
The covariance matrix of $X\in \RR^{3\times n}$ is given by 
$$\bar X {\bar X}^T \text{, where } \bar X=X-\frac{1}{n}1_n1_n^T. $$  
The covariance matrix is a symmetric positive semi-definite matrix and we denote its eigenvalues by $\lambda_1(X)\geq \lambda_2(X)\geq \lambda_3(X)\geq 0$ and define 
$$\eigcov(X)=(\lambda_1(X),\lambda_2(X),\lambda_3(X)) $$
We now define polynomials $p_2,p_4,p_6$ of degree $2,4,6$ which are invariant to permutations and rigid motions, by  
\begin{align*}
p_2(X)&=\Tr(\bar X {\bar X}^T)=\lambda_1(X)+\lambda_2(X)+\lambda_3(X)\\
p_4(X)&=\Tr(\left(\bar X {\bar X}^T\right)^2)=\lambda_1^2(X)+\lambda_2^2(X)+\lambda_3^2(X)\\
p_6(X)&=\Tr(\left(\bar X {\bar X}^T\right)^3)=\lambda_1^3(X)+\lambda_2^3(X)+\lambda_3^3(X)
\end{align*}
Since $p_2,p_4,p_6$ are invariant and of degree $\leq 6$ we can approximated them with our architecture $\gF(6,C)$ with $C$ large enough. 

It remains to show that there exists a continuous mapping $q:\RR^3\to \RR^3$ such that (dropping the dependence of the eigenvalues on $X$) $$q\left(\lambda_1+\lambda_2+\lambda_3,\lambda_1^2+\lambda_2^2+\lambda_3^2,\lambda_1^3+\lambda_2^3+\lambda_3^3\right)=(\lambda_1,\lambda_2,\lambda_3)$$
This follows from the fact that the three polynomials $s_1,s_2,s_3:\RR^3\to \RR$
\begin{align*}
s_1(\lambda_1,\lambda_2,\lambda_3)&=\lambda_1+\lambda_2+\lambda_3\\
s_2(\lambda_1,\lambda_2,\lambda_3)&=\lambda_1^2+\lambda_2^2+\lambda_3^2\\
s_3(\lambda_1,\lambda_2,\lambda_3)&=\lambda_1^3+\lambda_2^3+\lambda_3^3
\end{align*}
are permutation invariant polynomials known as the power sum polynomials, which generate the ring of permutation invariant polynomials on $\RR^3$, and as such, the map $\vec{s}=(s_1,s_2,s_3)$ induces a homeomorphism of $\RR^3/S_3$ onto the image of $\vec{s}$ (\cite{gonzalez2003c}, Lemma 11.13) . Similarly, the sorting function $\mathrm{sort}:\RR^3\to \RR^3 $ is an injective permutation invariant mapping which induces a homeomorphism of $\RR^3/S_3$ onto its image. Thus the sets 
$$\{\vec{s}(\lambda_1,\lambda_2,\lambda_3)| \, (\lambda_1,\lambda_2,\lambda_3)\in \RR^3\} \text{ and } \{\mathrm{sort}(\lambda_1,\lambda_2,\lambda_3)| \, (\lambda_1,\lambda_2,\lambda_3)\in \RR^3\}$$
are homeomorphic, and we can choose $q$ to be a homeomorphism (in particular, $q$ is continuous). 
\end{proof}

\end{document}

%% file: math_commands.tex
\usepackage{amsmath,amsfonts,bm}

\def\eqref#1{equation~\ref{#1}}

\def\1{\bm{1}}

\DeclareMathAlphabet{\mathsfit}{\encodingdefault}{\sfdefault}{m}{sl}
\SetMathAlphabet{\mathsfit}{bold}{\encodingdefault}{\sfdefault}{bx}{n}

\def\gA{{\mathcal{A}}}

\def\gD{{\mathcal{D}}}

\def\gF{{\mathcal{F}}}

\def\gL{{\mathcal{L}}}

\def\gO{{\mathcal{O}}}

\def\gT{{\mathcal{T}}}

\DeclareMathOperator{\Tr}{Tr}